\documentclass{article}

\usepackage{microtype}
\usepackage{graphicx}
\usepackage{subfigure}
\usepackage{booktabs} 

\usepackage{hyperref}

\usepackage[accepted]{icml2025}

\usepackage{amsmath}
\usepackage{amssymb}
\usepackage{mathtools}
\usepackage{amsthm,bm}

\usepackage[capitalize,noabbrev]{cleveref}

\theoremstyle{plain}

\theoremstyle{definition}

\theoremstyle{remark}

\usepackage[textsize=tiny]{todonotes}

\DeclareMathOperator*{\argmin}{arg\,min}

\icmltitlerunning{Solving Functional Optimization with Deep Networks and Variational Principles}

\begin{document}

\twocolumn[
\icmltitle{Solving Functional Optimization with Deep Networks and Variational Principles}



\icmlsetsymbol{equal}{*}

\begin{icmlauthorlist}
\icmlauthor{Kawisorn Kamtue}{yyy}
\icmlauthor{Jose M.F. Moura}{yyy}
\icmlauthor{Orathai Sangpetch}{zzz}
\end{icmlauthorlist}

\icmlaffiliation{yyy}{Department of Electrical and Computer Engineering, Carnegie Mellon University, Pittsburgh, USA}
\icmlaffiliation{zzz}{Department of Electrical and Computer Engineering, CMKL University, Bangkok, Thailand}

\icmlcorrespondingauthor{Kawisorn Kamtue}{kkamtue@andrew.cmu.edu}

\icmlkeywords{Machine Learning, ICML}

\vskip 0.3in
]



\printAffiliationsAndNotice{}

\begin{abstract}
Can neural networks solve math problems using first a principle alone? This paper shows how to leverage the fundamental theorem of the calculus of variations to design deep neural networks to solve functional optimization without requiring training data (e.g., ground-truth optimal solutions). Our approach is particularly crucial when the solution is a function defined over an unknown interval or support\textemdash such as in minimum-time control problems. By incorporating the necessary conditions satisfied by the optimal function solution, as derived from the calculus of variation, in the design of the deep architecture, CalVNet leverages overparameterized neural networks to learn these optimal functions directly. We validate CalVNet by showing that, without relying on ground-truth data and simply incorporating first principles, it successfully derives the Kalman filter for linear filtering, the bang-bang optimal control for minimum-time problems, and finds geodesics on manifolds. Our results demonstrate that CalVNet can be trained in an unsupervised manner, without relying on ground-truth data, establishing a promising framework for addressing general, potentially unsolved functional optimization problems that still lack analytical solutions.
\end{abstract}

\section{Introduction}
\label{intro}
The fundamental theorem of the calculus of variations is a powerful principle at the core of physics, underpinning classical mechanics, electromagnetism, and quantum theory. Nature inherently follows optimization principles, such as the principle of least action to dictate how light travels. This variational framework provides a powerful foundation for deriving optimality conditions. Given this prior knowledge, how can we incorporate this fundamental principle to deep neural networks to learn optimal solutions? 

We consider a general functional optimization problem where the support of the functions, defined over the interval $[0,t_f]$, is unknown. Specifically, the terminal time $t_f$ itself is subject to optimization, and the terminal state is constrained to lie within a predefined set. This formulation introduces significant challenges: traditional methods such as the forward methods \cite{neuralode,aipontryagin} and shooting methods \citep{QUARTAPELLE1990314,BONNANS2013281} are inapplicable, as they rely on known or fixed time horizons. Moreover, performance metrics are only valid for admissible trajectories\textemdash those that successfully reach the desired terminal state. Methods that directly minimizes the performance functional \cite{MOWLAVI2023111731} are not valid.

In this paper, we draw inspiration from the calculus of variations\textemdash a field dedicated to finding extrema of functionals through variations\textemdash to design a neural network based this first principle. Calculus of variations, the mathematics of optimizing functionals, optimizes directly over the function space, which contrasts with approaches that convert the function optimization into a parameter optimization by first parametrizing the possible function solutions, e.g., by representing the functions through splines. Our approach aims to solve functional optimization problems arising in engineering, optimal control, and differential geometry. We begin by formulating a variational framework for these problems, using the calculus of variations to derive the optimality conditions that make the functional variation vanish. While traditional methods solve these conditions analytically or numerically, they become intractable for nonlinear, second-order differential equations with complex boundary conditions. Instead, we propose a neural network that learns the optimal solution by directly minimizing these variations.

Additionally, the presence of extra constraints, such as boundedness or compactness requirements on the state function (e.g., finding curve on a manifold) and control functions, further complicates the problem. These constraints restrict the space of admissible control functions, often leading to optimal solutions that are discontinuous (resembling step functions) or undefined in certain regions. Such characteristics pose substantial difficulties for neural network training, frequently resulting in vanishing or exploding gradients. To date, no prior work has successfully addressed these challenges in their entirety.

This paper presents a method to integrate prior knowledge from calculus of variations, functional optimization, and classical control into the architectural design of deep models. We incorporate dynamical constraints, control constraints, and optimality conditions derived from the first principle into the loss function for training neural networks, enabling unsupervised learning. Our contributions are as follows.

\textbf{Main contributions}:
\begin{itemize}
    \item We provide a general framework of functional optimization by incorporating the fundamental theorem of calculus: train a deep neural network that makes the functional variation zero for all admissible variations
    \item Propose learning paradigms that effectively train CalVNet to derive the optimal solution.
    \item We use CalVNet to solve three optimization problems: one with constraints on control and one with constraints on state.
    \item Show that our CalVNet replicates the design of the Kalman filter, derives the bang-bang control, learns shortest curves (geodesics) on manifolds.
\end{itemize}
\section{Theory}
\subsection{Problem setting}
We present a context of functional optimization problem. Given an initial value problem, specified by a dynamical system and its initial condition
\begin{equation}
\label{eq:dynamic_system}
\begin{aligned}
    \dot{x}(t)&=f(x(t),u(t))\\
    x(0) &= x_0
\end{aligned}
\end{equation}
where $x:\mathbb{R}_{\geq 0}\mapsto \mathcal{X}\subseteq \mathbb{R}^{m}$ is the state function, $u:\mathbb{R}_{\geq 0}\mapsto \mathcal{U}\subseteq\mathbb{R}^n$ is the control function (if exists), and $f:\mathcal{X}\times\mathcal{U}\mapsto\mathcal{X}$ is a known function representing the dynamics. We suppose that $x$ is differentiable and $f$ is differentiable with respect to each variable. Unlike previous works that consider fixed support \citep{MOWLAVI2023111731} or fixed terminal state \citep{ponn}, we consider a more general stopping set $\mathcal{S}=\{(x(t),t)|\,\,\varphi(x(t),t)=0\} =\mathcal{X}\times \mathbb{R}_{\geq 0}$ where $s:\mathbb{R}^m\times\mathbb{R}_{\geq \bm 0} \mapsto \mathbb{R}^k $ is a differentiable function. This definition of $\mathcal{S}$ allows us to solve general functional optimization problems when the terminal state and time are not explicitly specified, e.g., finding the distance between two curves or finding the minimum time to reach the surface of a manifold. In these cases, we do not know the terminal point and terminal time beforehand.

Functional optimization problems involve finding a valid a trajectory  $x^\star:[0,t_f]\mapsto \mathcal{X}$ such that it reaches the terminal state $(x^\star(t_f),t_f)\in \mathcal{S}$ and minimizes some functional measure $\mathcal{L}(x,u)$ of the form
\begin{equation}
    \label{eq:functional_cost}
    \mathcal{L}(x,u) = q_T(x(t_f),t_f)+\displaystyle \int\limits_{0}^{t_f}g(x(t),\dot{x}(t),u(t))dt
\end{equation}
where $q_{T}$ is the terminal cost and~$g$ is the running cost. Not all pairs of functions $(x,u)$ are admissible trajectories since trajectories must satisfy a dynamical constraint $\dot{x}(t)=f(x(t),u(t))$ and $(x(t_f),t_f)\in \mathcal{S}$. The domain of integration $[0,t_f]$ can be variable, depending on each admissible control. The optimal control problem is therefore the constrained optimization 
\begin{equation}
\begin{aligned}
\min_{x,u} \quad &\mathcal{L}(x,u)\\
\textrm{s.t.} \quad &\dot{x}(t)=f(x(t),u(t)), \forall t\in [0,t_f] \\
&x(0)=x_0\\
&(x(t_f),t_f)\in\mathcal{S}
\end{aligned}
\label{eq:constrained_optimization}
\end{equation}

In~\eqref{eq:constrained_optimization}, the optimization variables are functions over variable support, say $\left\{u(t), t\in [0\,,\,t_f]\right\}$, where $t_f$ may be fixed or is to be optimized itself (like in the minimum time problem).

To handle dynamics constraints, the  Langragian multipliers function $\lambda(t)$ and the Lagrangian multiplier scalar $\lambda_f$ are introduced and the new functional becomes
\begin{equation}
    \begin{aligned}
    \label{eq:lagrangian}
    \mathcal{J} (x,u,\lambda,\lambda_f)  &= q_T(x(t_f),t_f)+ \lambda_f \varphi(x(t_f),t_f) \\
    &+\displaystyle \int\limits_{0}^{t_f}g(x,\dot{x},u)+ \lambda^T(f(x,u)-\dot{x}) dt
\end{aligned}
\end{equation}

For all admissible trajectories $(x,u)$, i.e. $(x,u)$ satisfying the dynamics $\dot{x}=f(x,u)$, we have $\mathcal{J}(x,u,\lambda)=\mathcal{L}(x,u)$. Therefore, the admissible optimal solution for~\eqref{eq:lagrangian} is also the optimal solution for \eqref{eq:constrained_optimization}.

\subsection{Calculus of variations}

Calculus of variations enables us to identify the optimal functions $s=(x,u,\lambda)$ that minimize $\mathcal{J}$. The fundamental theorem of calculus of variations states that  variation at the optimal solution is $0$,
$$\delta \mathcal{J}(s^\star,\delta s)=0,\quad \textrm{for all admissible $\delta s$}$$ 

we can use this powerful law to derive the necessary conditions at the optimal solution $s^\star=(x^\star,u^\star,\lambda^\star,\lambda^\star_f)$ 

To compute a variation $\delta{\mathcal{J}}$ at $(x,\lambda,u)$, we first add a perturbation $\delta s=(\delta x,\delta u,\delta \lambda)$. Note that $\delta s$ may not be arbitrary, i.e. there is a set of admissible $\delta s$ that makes $s+\delta s$ a valid trajectory. This perturbation causes the new trajectory to reach terminal state and time $(x_f+\delta x_f,t_f+\delta t_f)$

Then we compute $\Delta \mathcal{J} = \mathcal{J}(s+\delta s)-\mathcal{J}(s)$. Then $\delta \mathcal{J}(s,\delta s)$ is the first order terms in Taylor expansion of $\Delta J$, i.e. the linear terms of $\delta s$.


Using the fundamental theorem of calculus of variations, we can derive a set of necessary conditions $\{\psi_i\}_{i\in I}$ that makes $\delta \mathcal{J}(s^\star,\delta s)=0$ for all admissible $\delta s$

\begin{equation}
    \label{eq:necessary_conditions}
    \psi_i(s^\star,x^\star_f,t^\star_f)=0
\end{equation}

$\{\psi_i\}_{i\in I}$ consists of a system of partial differential, usually containing the Euler-Lagrange equation. This system of differential equation is generally nonlinear, time-varying, second-order, and hard-to-solve. Numerical methods also pose challenges due to the split boundary conditions--neither the initial values $(x(0),\dot{x}(0))$ nor the final values $(\lambda(t_f), \dot{\lambda}(t_f))$ are fully known.

\subsection{CalVNet: Calculus of Variations informed Network}

Instead of solving $\cref{eq:necessary_conditions}$ analytically or numerically, we propose leveraging neural networks' well-known capability as universal function approximators \citep{universal_approx} to learn $\left\{x(t), u(t), \lambda(t), t\in[0\,,\,t_f]\right\}$, along with the learnable parameter time interval $t_f$, that satisfy $\cref{eq:necessary_conditions}$. Unlike other parametric functions like spline functions \citep{DBLP:journals/corr/abs-2106-04315,software:stochman} and constrained expressions \citep{math5040057}, deep networks are overparameterized that can model even step functions (see results on \cref{sec:bangbang}). In the training stage, rather than directly matching the CalVNet's outputs to ground truth data $\left\{x(t)^\star, u(t)^\star, \lambda(t)^\star, t\in[0\,,\,t_f]\right\}$, our CalVNet \textbf{learns} to predict solutions that adhere to the $\cref{eq:necessary_conditions}$ by instead minimizing a smooth function
$$\Psi = \sum\limits_{i\in I}\|\psi_i\|_{d_i}^2$$
where $\|\cdot\|_{d_i}$ is an appropriate norm chosen.
Because this process incorporates the fundamental principle that variation vanishes at optimal solution, we interpret it as bringing to the neural networks ``prior knowledge''\citep{BETTI201683}. Our approach introduces an inductive bias into the CalVNet, allowing it to learn the optimal solution in an unsupervised manner. By simultaneously predicting both the state and the control, CalVNet eliminates the need for integration and can address optimal control problems with unknown terminal time.

The algorithm is detailed in \cref{alg:calvnet}. During the forward pass, CalVNet takes time as input and predicts the state $x_\theta(t)$, the control $u_\theta(t)$, and the costate $\lambda_\theta(t)$. The loss $\Psi$ is then calculated based on these predictions. By leveraging the automatic differentiation capabilities of neural networks \citep{ad,lu2021deepxde}, we can efficiently compute the derivatives and partial derivatives present in $\Psi$ by computing in-graph gradients of the relevant output nodes with respect to their corresponding inputs. In the experiments in \cref{sec:kalman_filter} and \cref{sec:bangbang}, we also incorporate additional architectural features into our CalVNet to enforce hard constraints and to allow CalVNet to learn even when the terminal time is unknown.

\begin{algorithm}[tb]
   \caption{CalVNet for functional optimization problem}
   \label{alg:calvnet}
\begin{algorithmic}
   \STATE {\bfseries Input:} Functional cost $J$, $\{\psi_i\}_{i\in I}$ 
   \STATE \qquad network parameter $\theta$, start time $t_0$, initial value $x_0$
   \STATE \qquad final time $t_f$
   \STATE {\bfseries Learnable parameters:} $\Phi=\{\theta,t_f,\lambda_f\}$
   \WHILE{unconverged}
   \STATE Sample time $\{t_k\}^{N}_{k=1}$ uniformly
   \STATE Compute $(x_\theta(t_k),\lambda_\theta(t_k),u_\theta(t_k))$ for all $k=1,..,N$
   \STATE Compute $\psi_i(x_\theta,\lambda_\theta,u_\theta)$ that makes $\delta J$ zero
   \STATE Compute $\Psi = \sum\limits_{i\in I}\|\psi_i\|_{d_i}^2$
   \STATE $\textrm{loss}=\textrm{MSELoss}(x_\theta(t_0),x_0)$ + $\Psi$
   \STATE $\Phi \leftarrow \Phi-\eta\nabla_{\Phi} \textrm{loss}$
   \ENDWHILE
   \STATE {\bfseries return} $\Phi$
\end{algorithmic}
\end{algorithm}
\section{Designing the optimal linear filter}
\label{sec:kalman_filter}
In this section, our goal is to design a linear filter that provides the best estimate of the current state based on noisy observations. The optimal solution is known as ``Kalman Filtering'' \citep{Klmn1961NewRI}, which is one of the most practical and computationally efficient methods for solving estimation, tracking, and prediction problems. The Kalman filter has been widely applied in various fields from satellite data assimilation in physical oceanography, to econometric studies, or to aerospace-related challenges \citep{Leonard1985DiscoveryOT,6400245}. The optimal solution being known, the Kalman filter is the ground truth that serves to benchmark CalVNet.
\subsection{Kalman Filter}
 Reference \citet{athans1967} formulated a variational approach to derive the Kalman filter as an optimal control problem. We consider the dynamical system
\begin{equation}
    \label{kalman}
    \begin{aligned}
        \dot{x}(t) &= Ax(t)+Bw(t), \:\: 0\leq t\leq t_f,\:\:\:
        \mbox{ $w_{t-1}\sim \mathcal{N}(0,\textbf{Q})$}\\
        y(t) &= Cx(t) + v(t), 
        \mbox{ $v_{t-1}\sim \mathcal{N}(0,\textbf{R})$}\\
        x(0) &\sim\mathcal{N}(x_0,\Sigma_0)
    \end{aligned}
\end{equation}
where $x(t)\in\mathbb{R}^n$ is the state, $y(t)\in\mathbb{R}^m$ is the observation. $A\in\mathbb{R}^{n\times n}$ is the state transition matrix, $B\in\mathbb{R}^{n\times r}$ is the input matrix, and $C\in\mathbb{R}^{n\times r}$ is the measurement matrix. The white Gaussian noise $w(t)$ (resp. $v(t)$) is the process (resp. measurement) with covariance $\textbf{Q}$ (resp. $\textbf{R}$) noise. We assume that $x(0),w(t),v(t)$, are independent of each other.  Kalman designed a recursive filter that estimates the state by
\begin{equation}
    \label{eq:kalman_gain}
    \begin{aligned}
        \dot{\hat{x}}(t) &= A\hat{x}(t)+G(t)\Big[Cy(t) - A\hat{x}(t)\Big]\\
    \hat{x}(0) &= x_0
    \end{aligned}
\end{equation}
where $G(t)$ is the Kalman gain to be determined. Given the state estimation $\hat{x}(t)$ at time $t$, the error covariance defined as $$\Sigma(t)= \mathbb{E}\Big[(\hat{x}-x)(\hat{x}-x)^T\Big]$$ has the following dynamics
\begin{equation}
\begin{split}
    \label{eq:kalman_dynamics}
    \dot{\Sigma}(t) = &\Big[A-G(t)C\Big]\Sigma(t) +\Sigma(t)\Big[A-G(t)C\Big]^T\\
    &+ BQB^T+G(t)RG(t)^T\\
    \Sigma(0)= &\Sigma_0
\end{split}
\end{equation}
where $\Sigma(t)$ is the $n\times n$ error covariance matrix. The goal of Kalman filter is to find the optimal gain (perceived in this variational approach as a control) $G^\star(t)$ such that the final cost
$$q_T(\Sigma(T)) = \textrm{tr}\left[\Sigma(T)\right]$$ is minimized, or equivalently the $L_2$ norm between the estimation and the actual state is minimized. In this case, the stopping set is $\mathcal{S}=\{(\Sigma(t),t) |\, t=T\}$, there is no constraint on terminal state and terminal time is fixed at $T$.
\begin{equation}
    \begin{aligned}
    \label{eq:kalman_functional}
    \mathcal{J} (\Sigma,G,\lambda)  &= \textrm{tr}\left[\Sigma(T)\right]+\displaystyle \int\limits_{0}^{t_f}\lambda^T(f(x,u)-\dot{x}) dt
\end{aligned}
\end{equation}
where $f$ is the dynamics described in \cref{eq:kalman_dynamics}. Deriving $\delta \mathcal{J}$ from \cref{eq:kalman_dynamics} (see \cref{appendix:kalman}), the necessary conditions to solve for the optimal Kalman gain are
\begin{equation}
    \label{eq:solution_kalman}
    \begin{split}
\psi_1 &= \dot{\Sigma}^\star-f(\Sigma^\star,G^\star) = 0\\
\psi_2 &= \displaystyle \dot{\lambda^{\star}}^T+\frac{\partial \mathcal{H}}{\partial \Sigma}\Big|_\star = 0\\
        \psi_3 &=\displaystyle \frac{\partial \mathcal{H}}{\partial G}\Big|_{\star} = 0 \\
    \psi_4 &= \lambda^\star(T)^T - \bm I_n = 0\\
    \end{split}
\end{equation}
where the Hamiltonian $\mathcal{H}=\textrm{tr}\left[\lambda^Tf(\Sigma^\star,G^\star)\right]$
\subsection{Learning the Kalman Filter with CalVNet}\label{subsec:kalmanfilter}


\textbf{Architecture}: The state, costate, and control estimators are modeled by 6-layer feedforward neural networks with hyperbolic tangent activation. Since $\Sigma$ is both symmetric and positive semi-definite, we embed this inductive bias into our neural network architecture. Specifically, the state estimator outputs an intermediate matrix $P$ and estimates the error covariance $\Sigma$ as $\Sigma = P^T P$, ensuring symmetry and positive semi-definiteness. We adopt the feedback loop design in engineering so that the control estimator only takes the output state as input.  

\textbf{Training}: We adopt curriculum training, as optimizing loss with multiple soft constraints can be challenging \citep{krishnapriyan2021characterizing}. We set the loss function to be 
\begin{equation*}
    \textrm{Loss}_{\theta} = \|\Sigma(0)-\Sigma_0\|^2_2 + \alpha \Psi
\end{equation*}
 During each epoch, 5000 points are uniformly sampled from time $[0\,,\,T]$. After every $5000$ epochs, we increment the value of $\alpha$ by a factor of $1.04$. All neural networks are initialized with Glorot uniform initialization \citep{Glorot2010UnderstandingTD}. We train CalVNet using stochastic gradient descent with the initial learning rate $8\times 10^{-4}$.

\textbf{Evaluation}: For a fair evaluation, we take the estimated $G$ from CalVNet and use the fourth-order Runge-Kutta integrator \citep{Runge1895} in $\mathtt{scipy.integrate.solve\_ivp}$ to derive the trajectory of the state. This is necessary because the  state estimated by CalVNet might not adhere to the dynamics constraints, making it into an implausible trajectory.
\subsection{Results}
For our experiment, we set $$A\!=\!\begin{bmatrix}
    \bm 0 & \bm I_2\\
    \bm 0& \bm 0
\end{bmatrix}\!\in\!\mathbb{R}^{4\times 4},B=\begin{bmatrix}
    \bm 0 \\
    \bm I_2
\end{bmatrix}\!\in\!\mathbb{R}^{4\times 2}, C=\bm I_4, T=5.0$$

\begin{figure*}[htp!]  \includegraphics[width=\textwidth,height=4cm]{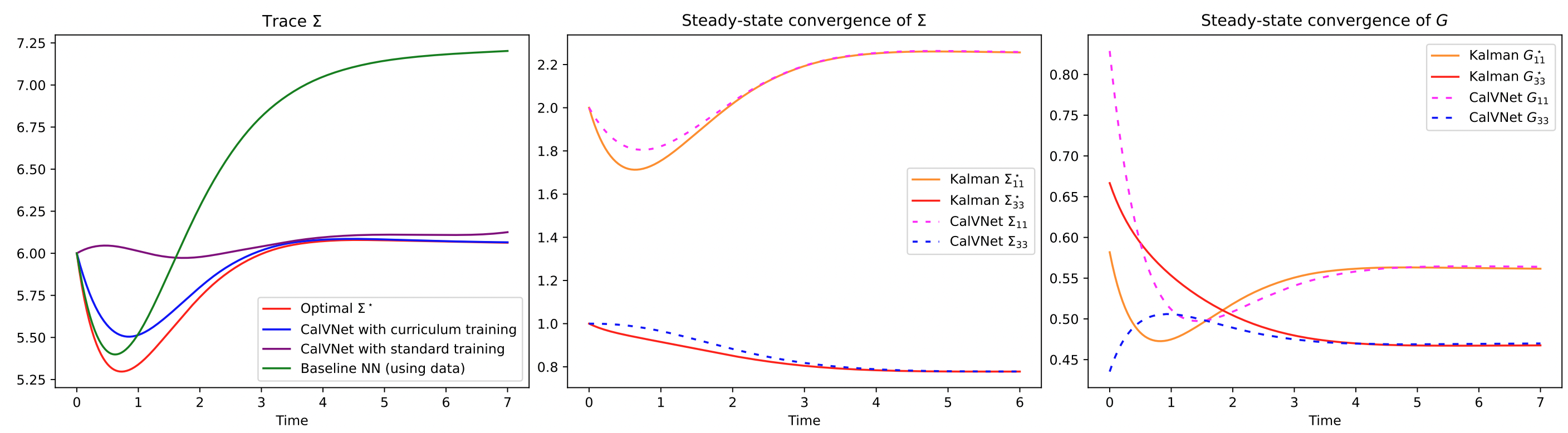}
  \caption{CalVNet learns the Kalman filter, deriving the optimal value of the functional cost. The baseline NN performs well in the time interval where ground truth is available, but fails to learn the optimal steady-state Kalman gain $G^\star_\infty$, resulting in diverging error. The baseline PINN shows diverging error. CalVNet learns the optimal steady-state error covariance $\Sigma^\star_\infty$ and Kalman gain $G^\star_\infty$ and they remain convergent beyond the time interval of the problem $[0,5]$}
  \label{fig:kalman_result}
\end{figure*}
This dynamical system models a kinematics system where the state $x$ corresponds to position and velocity and the control $u$ corresponds to the force applied to the state. With these experimental settings, Kalman filtering reaches a steady state where $\Sigma^\star$ converges (hence, the Kalman gain converges to $G^\star_\infty$). We compare our method against two baselines: 1) the baseline NN trained with 50 points of ground truth control $G^\star$ sampled from the time interval $[0\,,\,2.0]$, covering the transient phase of the Kalman filter before it reaches steady-state and 2) the baseline PINN that enforces the dynamics constraints and directly minimize the cost functional $q_T$ \citep{MOWLAVI2023111731} instead of variations.
We evaluate and compare the trace of $\Sigma$ generated by CalVNet, the baseline methods, and the optimal Kalman gain. \cref{fig:kalman_result} shows that, even though there is some discrepancy between the CalVNet's control output $G$ and the Kalman gain $G^\star$ during the transient phase, CalVNet matches the optimal Kalman gain $G^\star_\infty$ at the terminal time, while the baseline diverges. \cref{table:kalman_result} shows that the baseline PINN that learns to satisfy dynamics constraint and to minimize the cost functional without using optimality conditions shows a divergent behavior. This result demonstrates that directly optimizing functional measure (e.g. $q_T$) can be more unstable that optimizing variations. CalVNet's trajectory of $(\Sigma,G)$ converges to their corresponding optimal values $(\Sigma^\star_\infty,G^\star_\infty)$. Since the gain $G$ is learned as a function of one input $\Sigma$, 
 $(\Sigma,G)$ remains convergent even after time interval of the problem $[0,5]$, allowing us to use CalVNet in different time horizon. CalVNet learns the correct relationship between $\Sigma^\star_\infty$ and $G^\star_\infty$, equivalent to deriving the Riccati equation. The error covariance, and $\textrm{tr}(\Sigma)$ remain close to the ground truth of the analytical solution beyond time $T=5$. The discrepancy during the transient time does not affect the overall performance, since CalVNet's control $G$ converges to the optimal steady-state value $G^\star_\infty$. In practice, this is what usually matters, since in Kalman filter practice, the steady-state $G^\star_\infty$ is often pre-computed and used instead of $G^\star(t)$.
\begin{table}[t]
\caption{Trace and Convergence of predicted $\Sigma$ by CalVNet and baseline PINN}
\label{table:kalman_result}
\vskip 0.15in
\begin{center}
\begin{small}
\begin{sc}
\begin{tabular}{lcr}
\toprule
Method & tr($\Sigma$) & Convergence  \\
\midrule
Optimal Gain (Kalman)    & 6.06 & $\surd$ \\
CalVNet & 6.07& $\surd$\\
Baseline PINN & 14.7 &$\times$\\
\bottomrule
\end{tabular}
\end{sc}
\end{small}
\end{center}
\vskip -0.1in
\end{table}
We investigated the effect of using curriculum training. As shown in \cref{fig:kalman_result}, using curriculum training results in a trajectory with a smaller trace of the error covariance throughout the interval of interest, especially during the transient phase. 

\section{Learning the Minimum Time optimal control}
\label{sec:bangbang}
In this section, we seek the optimal control strategy that drives a state from an arbitrary initial position to a specified terminal position in the shortest possible time. In practice, the control is subject to constraints, such as maximum output levels. The optimal control strategy for the minimum time problem is commonly known as ``bang-bang'' control. Examples of bang-bang control applications include guiding a rocket to the moon in the shortest time possible while adhering to acceleration constraints \citep{optimalbook}.
\subsection{The Minimum time problem}
We illustrate the CalVNet with the following problem. Consider the kinematics system 
\begin{equation}
\label{eq:bangbang_dynamics}
    \begin{bmatrix}
        \dot{x}_1(t) \\ \dot{x}_2(t)
    \end{bmatrix}=\begin{bmatrix}
        0&1\\
        0&0
    \end{bmatrix}\begin{bmatrix}
        x_1(t) \\ x_2(t)
    \end{bmatrix}+\begin{bmatrix}
        0 \\ 1
    \end{bmatrix}u 
\end{equation} 
where $x_1,x_2,u$ correspond to the position, velocity, and acceleration of a mobile platform. The goal is to drive the system from the initial state $(x_1(0),x_2(0))=(p_0,v_0)$ to a final destination $\mathcal{S}=\{(x(t),t) \;| \;\|x(t)\|= 0\}$. where $x(t)\in\mathbb{R}^n$ is the state at time $t$, $u(t)\in\mathbb{R}^m$ is the control at time $t$. We are interested in finding the optimal control $\left\{u^\star(t), t\in[0\,,\,t^\star_f]\right\}$ that drives the state from $x_0$ to $x_f$ in a minimum time $t^\star_f$. The performance measure can be written as
\begin{equation}
    \label{eq:minimum_time_functional}
    T(x,u) = \displaystyle \int\limits_{0}^{t_f}1dt
\end{equation}
where $t_f$ is the time in which the sequence $(x,u)$ reaches the terminal state. Note that here $t_f$ is a function of $(x,u)$ since the time to reach the target state depends on the state and control.
In practice, the control components may be constrained by requirements such as a maximum acceleration or maximum thrust
\begin{equation*}
    \label{constraint_u}
    \begin{aligned}
        |u_i(t)| \leq 1, \quad i\in[1,m] \quad t\in[t_0,t_f]
    \end{aligned}
\end{equation*}
where $u_i$ is the $i$th component of $u$. In the case of control with constraint, the variation can be nonzero when the optimal solution is at the boundary.
\begin{equation}
\mathcal{J} (x,u,\lambda) = \lambda_f\psi(x_f)+ \int\limits_{0}^{t_f}1+\lambda^T(f(x,u)-\dot{x}) dt
\label{eq:bangbang_functional}
\end{equation}

Deriving $\delta \mathcal{J}$ gives us the necessary conditions at the valid optimal solution $(x^\star,u^\star,\lambda^\star)$ (see \cref{appendix:bangbang})
\begin{equation}
\label{eq:solution_bangbang}
    \begin{split}
    \psi_1 &= \dot{x}^\star-f(x^\star,u^\star) =0\\
    \psi_2 &=\dot{\lambda^\star} - \begin{bmatrix}
        0\\
        -\lambda_1^\star
    \end{bmatrix} = \bm 0\\
    \psi_3 &=u^\star -\argmin_u  \mathcal{H}(x^\star,u,\lambda^\star) = 0\\
    \psi_4 &= 1+ \lambda_2(t^\star_f)u(t^\star_f) = 0\\
    \end{split}
\end{equation}
where $f$ is the dynamic function described in \cref{eq:bangbang_dynamics} and the Hamilton $\mathcal{H}(x,u,\lambda)=1+\lambda_1x_2+\lambda_2u$. 
 \subsection{Learning bang-bang control with CalVNet}\label{subsec:bangbang}
 In our approach, the state estimator, costate estimator, and control estimator are modeled by 6-layer feedforward networks. The control estimator has the hyperbolic activation at the final output to ensure the control is bounded by $1$. The learnable parameter $t_f$ is subjected to the constraint $x(t_f)=x_f$.

\begin{figure*}  \includegraphics[width=\textwidth,height=4cm]{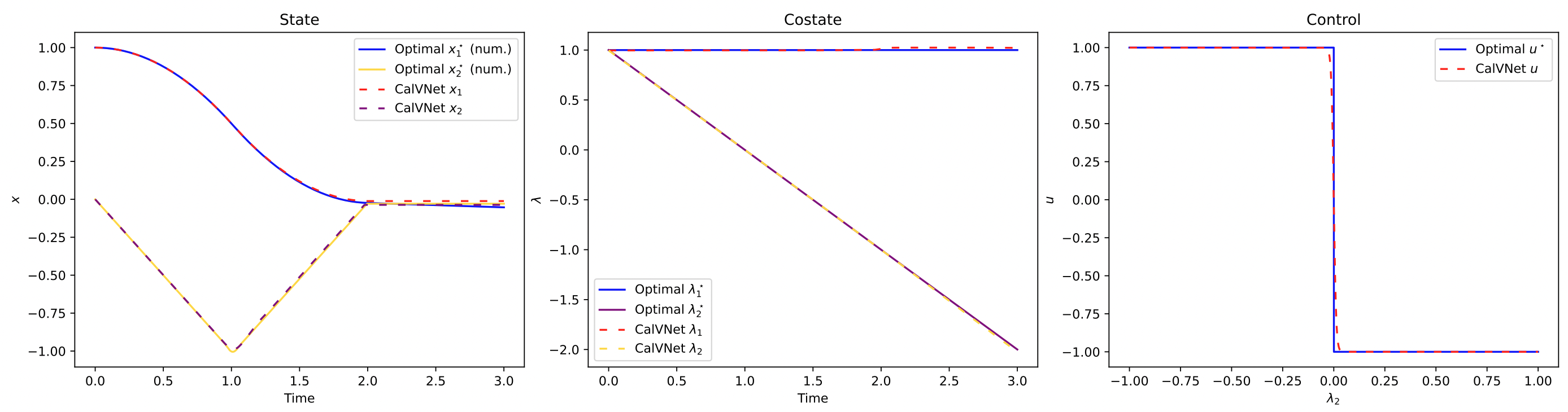}
  \caption{CalVNet generates the trajectory of the state, the costate, and the control over the time interval of interest that matches the optimal trajectory. Most importantly, CalVNet learns the bang-bang behavior where control $u$ is a negative sign function of $\lambda_2$ and correctly learns the minimum time $t^\star_f$.}
  \label{fig:bangbang_result}
\end{figure*}

\textbf{Training}: We propose a new paradigm for training CalVNet for minimum-time problems. First, we set a time $T$ that is sufficiently larger than $t^\star_f$. We start by pretraining the costate estimator such that the costate estimator is not a zero function (see Appendix~\ref{appendix:bangbang}). Secondly, we propose sequential and alternate training. \cref{eq:solution_bangbang} suggests that the optimal control $u^\star$ as a function of $(x^\star,\lambda^\star)$ can be learned without knowing $(x^\star,\lambda^\star)$. Therefore, in the first step, we can generate a random $(x,\lambda)$ and train the control estimator to minimize $\mathcal{H}(x,\lambda,u)$. We freeze the state and costate estimator and take $n$ gradient update for the control estimator since $u^\star =\argmin_u \mathcal{H}(x,\lambda,u)$. Next, we freeze the control estimator and train the state and costate estimator by uniformly sampling 5000 points from time interval $[0,T]$ and perform one gradient update for the state and costate estimator before going back to the first step again. This can prevent vanishing gradients or exploding gradients. We also compute the gradient of the loss function with respect to the variable $t_f$, allowing it to be optimized during backpropagation. We train CalVNet using stochastic gradient descent with the initial learning rate $8\times 10^{-4}$.

\textbf{Evaluation}: Similar to the experiment in \cref{subsec:kalmanfilter}, we generate the control estimate from CalVNet and use a fourth-order Runge-Kutta integrator to estimate the state trajectory. For the baseline, we employ the optimal (bang-bang) control and integrate it with the fourth-order Runge-Kutta method. During prediction, we consider the state to have reached the target if the Euclidean distance between them is less than $\epsilon=0.05$.
\subsection{Results}
For our experiment, we set $x_0 = \begin{bmatrix}
    1\\0
\end{bmatrix},x_f=\begin{bmatrix}
    0\\0
\end{bmatrix}$, $T=3.0$. The optimal control is to apply the acceleration $-1$ from time $[0,1]$ and acceleration $+1$ from time $]1,2]$ that will drive the state from the initial state $x_0$ to the target state $x_f$ in minimum time $t^\star_f=2$ seconds. The control switches from $-1$ to $+1$ at the switching time at $t=1$ where $\lambda^\star_2(t)=0$ as shown in \cref{fig:bangbang_result}.

\cref{fig:bangbang_result} show that the generated trajectory of state and costate match the optimal solution. CalVNet learns a control strategy that exhibits ``bang-bang" behavior, switching from $+1$ to $-1$ when $\lambda_2$ changes sign. Since standard neural networks inherently produce continuous functions, there is a small discrepancy between the predicted control and the theoretical bang-bang control. This limitation may, in fact, better reflect real-world scenarios, as the control cannot switch instantaneously between two extremes. While reducing this discrepancy is possible by using a larger control estimator and more computational resources to compute gradients of higher magnitude, such optimization is beyond the scope of this work. \cref{fig:bangbang_result} demonstrates that the trainable variable $t_f$ in CalVNet successfully converges to the true value of $t^\star_f = 2$. This key result highlights CalVNet's ability to learn when the terminal time is unknown.

\section{Geodesics on manifold}\label{sec:diff_geo}
In this section, we explore the task of finding the shortest curve on a manifold from a starting point to a stopping set $\mathcal{S}$. This general formulation encompasses a variety of specific problems, such as projecting a point in $\mathbb{R}^d$ onto a submanifold, determining a tangent curve from a point to a submanifold, or computing the shortest path between two points on a manifold. It is well-established that geodesic curves, which represent the shortest paths on a manifold, are the solutions to such problems. Beyond their theoretical importance, geodesics have significant practical applications. For instance, they are crucial in optimizing the transport of goods and passengers by minimizing time and energy costs. Additionally, they play a key role in numerical methods, where they help satisfy constraints during numerical optimization processes.
\subsection{Minimum length curve}
Given a Riemannian manifold $(\mathcal{M},g)$ where $\mathcal{M}$ is a smooth submanifold of $\mathbb{R}^n$ and $g$ is a Riemannian metric $g$ on $\mathcal{M}$ that assigns to each point $p\in\mathcal{M}$ a positive-definite inner product $g_p:T_p\mathcal{M}\times T_p\mathcal{M} \rightarrow \mathbb{R}$ where $T_p\mathcal{M}$ is a tangent space at $p$. For simplicity, we consider the usual $\|\cdot\|$ of $\mathbb{R}^n$. Given the curve $\gamma:[0,1]\rightarrow \mathcal{M}$, the arc length is defined as
\begin{equation*}
    L(\gamma) = \int\limits_{0}^{1} \|\dot{\gamma}(t)\| dt
\end{equation*}


The functional $L$ is not smooth and the minimizer is non-unique. Instead, we can define energy functional 
\begin{equation*}
    E(\gamma) = \int\limits_{0}^{1} \|\dot{\gamma}(t)\|^2 dt
\end{equation*}

The energy functional is locally uniformly convex, therefore the minimizer is unique. The optimal solution minimizing the energy functional also minimizes length functional (see Appendix).

Suppose the equation for the manifold $\mathcal{M}$ is described by $f(p)=0$ and the equation for stopping set $\mathcal{S}$ is $\varphi(p)=0$. We introduce lagrange multiplier and solve for the

\begin{equation}
\mathcal{J} (\gamma,\lambda,\lambda_f) = \lambda_f\varphi(x_f)+ \int\limits_{0}^{1}\|\dot{\gamma}\|^2 +\lambda(t)^Tf(\gamma(t))dt
\label{eq:energy_functional}
\end{equation}
Deriving $\delta J$ from \cref{eq:energy_functional} yields
\begin{equation}
\label{eq:solution_geodesics}
    \begin{aligned}
    \psi_1 &=f(\gamma^\star(t))=0 \qquad\quad \forall t\in[0,1]\\
    \psi_2 &= {\lambda^{\star}}^T\frac{\partial f}{\partial \gamma}-\ddot{\gamma}^\star = 0 \ \quad \forall t\in[0,1]\\
    \psi_3 &= \dot{v}(1) - \lambda_f\frac{\partial\varphi}{\partial \gamma}\\
    \gamma^\star(0) &= p_0 \\
    \varphi(\gamma^\star(1)) &= 0 \\ 
\end{aligned}
\end{equation}
The second equation $\psi_2$ in \cref{eq:solution_geodesics} is equivalent to the definition of "geodesics": the acceleration is perpendicular to the tangent plane (the covariant derivative of $\dot{\gamma}$ relative to $\dot{\gamma}$ is $0$. 
\subsection{Experiments}

\begin{figure}[ht]
\begin{center}
\centerline{\includegraphics[width=0.7\columnwidth]{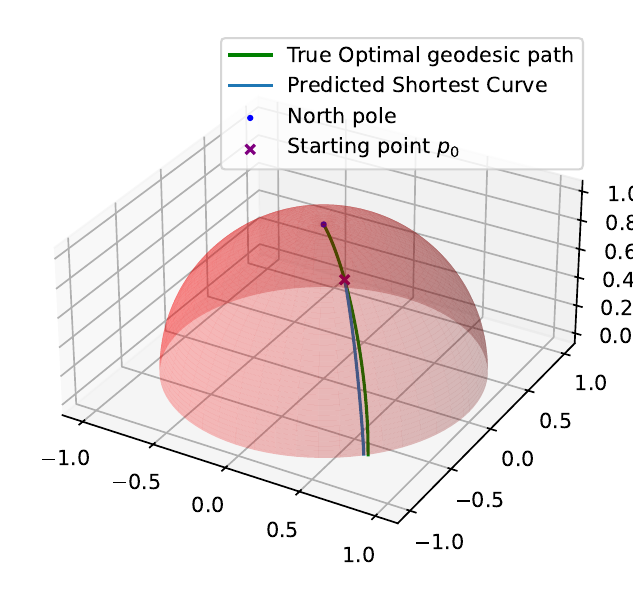}}
\caption{CalVNet finds the geodesics path from the point $p_0$ to the equator. The optimal path is given by a geodesics that past through the north pole and the point $p_0$}
\label{fig:sphere}
\end{center}
\vskip -0.2in
\end{figure}

\begin{figure}[ht]
\begin{center}
\centerline{\includegraphics[width=0.7\columnwidth]{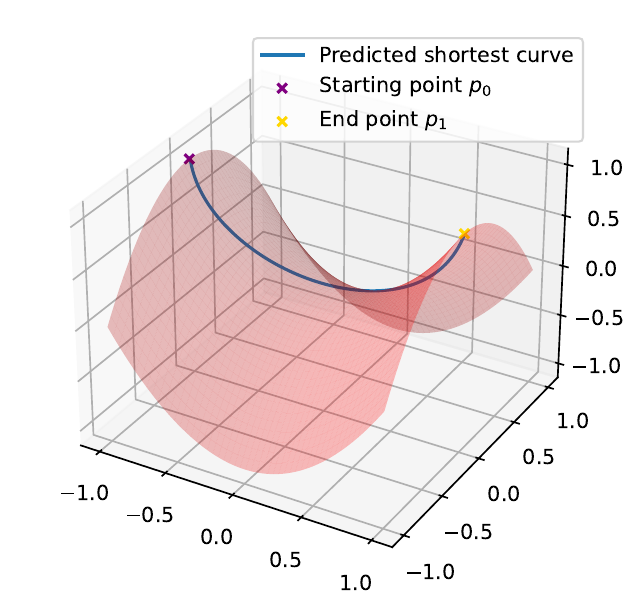}}
\caption{CalVNet finds the geodesics path on hyperbolic paraboloid from point $p_0$ to $p_1$}
\label{fig:hyperbolic_paraboloid}
\end{center}
\vskip -0.2in
\end{figure}

We consider two problems
\begin{itemize}
    \item finding the shortest path on a sphere $\mathbb{S}^2 = \{x^2+y^2+z^2=1\}$ from a point $p_0\in\mathbb{S}^2$ to the equator. The stopping set is defined by the equator $\varphi(x,y,z)=(x^2+y^2+z^2-1)^2+z$
    \item finding the shortest path on a hyperbolic paraboloid $\mathbb{H}^2 = \{z=x^2-y^2\}$ from a point $p_0\in\mathbb{H}^2$ to $p_1\in\mathbb{H}^2$
\end{itemize}
The results are shown in \cref{fig:sphere,fig:hyperbolic_paraboloid}. The optimal curves found lie on respective manifold and representing geodesics (acceleration is orthogonal to the tangent plane). For hyperbolic paraboloid, there is closed-form solution for the geodesics \citep{mariobook}. This motivates our work to use deep learning to learn geodesics on complex surfaces.

\section{Related Work}

Our approach aligns with the use of neural networks for solving optimal control problems and is inspired by existing literature on integrating constraints into neural network architectures. Below, we provide a concise overview of these areas, highlighting their relevance. We provide a brief overview of these areas and emphasize how our work distinguishes itself from them.

\textbf{Enforcing dynamics constraints in neural networks}:
Dynamics constraints in neural networks can be addressed through two main approaches: (1) designing specialized architectures that inherently satisfy the constraints (hard constraints), and (2) incorporating the constraints into the loss function, as done in Physics-Informed Neural Networks (PINNs) \cite{pinn} (soft constraints).

In hard constraint approaches, \citet{aipontryagin} enforce dynamic constraints using neural ODEs \citep{neuralode} to learn the optimal control. ODE-based methods primarily address the forward problem by integrating the state to the terminal time, calculating the loss function, and minimizing it. This framework is not applicable when the terminal time $t_f$ is unknown and must be optimized, or when the terminal state is prescribed. Similarly,  \citet{ponn} parameterize the state $x$ and express the control $u$ in terms of $x$ and its higher-order derivatives to satisfy the dynamic constraints. However, such a representation is not always feasible in general dynamics. 

In a soft constraint approach, \citet{MOWLAVI2023111731} employ PINNs to parameterize the state $x$ and control $u$, ensuring they satisfy the dynamics. The neural network weights are then updated to minimize the performance measure. However, this direct method assumes the performance metric can always be calculated—requiring the supports of the relevant functions to be fixed and known.

In contrast, our method uses the indirect method by leveraging the calculus of variations, enabling us to address cases where the terminal time and terminal state are variables (moving boundary). Our approach simultaneously learns the optimal control and the minimum time, even under these conditions.

\textbf{Incorporating optimality conditions in neural networks}: Several works have used optimality conditions of constrained optimization in neural networks. Reference \citet{amos2017optnet} and \citet{donti2021dc3} incorporate Karush–Kuhn–Tucker (KKT) conditions in implementing backward passes in neural networks. But this is constrained optimization over constant variables (parameters) while we optimize over functions with a dynamic constraint. Reference \citet{doi:10.1137/22M154209X} and \citet{betti1} propose using neural networks to parameterize the state and costate that learns to satisfy KKT and PMP conditions. However, these works only consider problems where the support is fixed. This approach can not be extended to a problem where the support is unknown, e.g., as in the minimum time problem. While \citet{ponn} considers learning the terminal time, their approach remains limited when the terminal state is not specified (e.g. when finding a projection onto manifolds).
\section{Conclusion}
We present a novel paradigm that integrates calculus of variations into neural networks for learning the solutions to functional optimization problems arising in many engineering and technology and scientific problems. Our CalVNet is unsupervised, generalizable and can be applied to a general functional optimization problems with moving boundaries that other related works have not addressed. We illustrate the CalVNet framework with classical problems of great applied significance and show that it successfully recovers the Kalman filter, bang-bang control solutions and geodesics. By leveraging the calculus of variations, we can analyze variations in the terminal state and time, and CalVNet successfully optimizes this variable in the minimum time problem\textemdash something most prior works fail to do. Although these solutions have been derived analytically in the past, we experiment with these problems, especially bang-bang control where no prior work has managed to use neural network to solve before, so that we can evaluate our results with the analytical optimal solutions. Our work paves the way for applying deep neural networks to more complex, higher-dimensional, and analytically intractable functional optimization problems.






\bibliography{main_paper}
\bibliographystyle{icml2025}

\newpage
\appendix
\onecolumn



\section{Calculus of variation and Pontryagin's maximum principle}
Suppose we want to find the control $u^\star(t),t\in[0,t_f]$ that causes the system 
\begin{equation}
\label{eq:appendix:dynamics_constraint}
\begin{split}
    \dot{x} &= f(x,u)\\
    x(0)&=x_0
\end{split}
\end{equation}
, where $f$ is a continuous function with continuous partial derivatives with respect to each variable, to follow an admissible trajectory $x^\star(t),t\in[0,t_f]$ that reaches the stopping set $\mathcal{S}$, i.e., $(x(t_f),t_f)\in \mathcal{S}$ and  minimizes the performance measure 
\begin{equation*}
    \mathcal{L}(x,u) = q_T(x_f,t_f) + \int_{0}^{t_f} g(x(t),u(t),t)dt
\end{equation*}
We consider the stopping set $\mathcal{S}$ to be of a general form $\mathcal{S}=\{(x(t),t)|\,\,s(x(t),t)=0\} =\mathcal{X}\times \mathbb{R}_{\geq 0}$ where $\varphi:\mathbb{R}^m\times\mathbb{R}_{\geq \bm 0}$ is a differentiable function. We suppose that the integrand $g$ and $q_T$ are smooth.
We introduce the (vector function) Lagrange multipliers $\lambda$, also known as costates. The primary function of $\lambda$ is to enable us to make perturbations $(\delta x, \delta u)$ to an admissible trajectory $(x,u)$ while ensuring the dynamic constraints in~\eqref{eq:appendix:dynamics_constraint} remain satisfied.
Suppose we have an admissible trajectory $(x,u,\lambda)$ such that reaches the terminal state $x_f$ at time $t_f$, the new functional $\mathcal{J}$ is 
    \begin{align*}
    \mathcal{J} (x,u,\lambda,x_f,t_f) &= q_T(x(t_f),t_f)+\lambda_f\varphi(x_f,t_f)+ \int\limits_{0}^{t_f}g(x,u)+ \lambda^T(f(x,u)-\dot{x}) dt\\
\end{align*}
Defining the Hamiltonian $\mathcal{H}=g(x(t),u(t))+\lambda(t)^Tf(x(t),u(t))$. The calculus of variations studies how making a small pertubation to $(x,u,\lambda)$ changes the performance. Suppose the new trajectory $(x+\delta x, u+\delta u,\lambda + \delta \lambda)$ reaches new terminal state $(x_f+\delta x_f,t_f+\delta t_f)$. Using Taylor expansion to the first order, the change in performance is
\begin{align*}
    \Delta\mathcal{J} &=\mathcal{J} (x+\delta x,u+\delta u,\lambda+\delta \lambda,x_f+\delta x_f,t_f+\delta t_f)-\mathcal{J} (x,u,\lambda,x_f,t_f)\\
    &=  \frac{\partial q_T}{\partial x}\delta x_f+\frac{\partial q_T}{\partial t}\delta t_f+ \lambda_f\frac{\partial \varphi}{\partial x}\delta x_f+\lambda_f\frac{\partial \varphi}{\partial t}\delta t_f+\delta\lambda_f\varphi(x_f,t_f)+\left[\mathcal{H}(x,u,\lambda,t_f)-\lambda(t_f)^T\dot{x}(t_f))\right]\delta t_f\\
    &\quad+\int\limits_{0}^{t_f}\frac{\partial \mathcal{H}}{\partial x}\delta x + \frac{\partial \mathcal{H}}{\partial u}\delta u + \left[\frac{\partial \mathcal{H}}{\partial \lambda}-\dot{x}\right]^T\delta \lambda - \lambda^T\delta \dot{x}dt+o(\|\delta x\|,\|\delta u\|,\|\delta \lambda\|,\|\delta t_f\|)\\
    &= \delta\lambda_f\varphi(x_f,t_f)+ \left[\frac{\partial q_T}{\partial x}+\lambda_f \frac{\partial \varphi}{\partial x}\right]\delta x_f+\left[\lambda_f\frac{\partial \varphi}{\partial t}+\frac{dq_T}{dt}(x_f,t_f)+\mathcal{H}(x,u,\lambda,t_f)-\lambda(t_f)^T\dot{x}(t_f))\right]\delta t_f\\
    &\quad+\int\limits_{0}^{t_f}\frac{\partial \mathcal{H}}{\partial x}\delta x + \frac{\partial \mathcal{H}}{\partial u}\delta u + \left(\frac{\partial \mathcal{H}}{\partial \lambda}-\dot{x}\right)^T\delta \lambda - \lambda^T\delta \dot{x}dt+o(\|\delta x\|,\|\delta u\|,\|\delta \lambda\|,\|\delta t_f\|)\\
   &= \delta\lambda_f\varphi(x_f,t_f)+ \left[\frac{\partial q_T}{\partial x}+ \lambda_f\frac{\partial \varphi}{\partial x}\right]\delta x_f+\left[ \lambda_f\frac{\partial \varphi}{\partial t} +\frac{\partial q_T}{\partial t}+\mathcal{H}(x,u,\lambda,t_f)-\lambda(t_f)^T\dot{x}(t_f))\right]\delta t_f - \lambda(t_f)^T\delta x(t_f)\\
   &+\int\limits_{0}^{t_f} \left[\dot{\lambda}+\frac{\partial \mathcal{H}}{\partial x}\right]\delta x + \frac{\partial \mathcal{H}}{\partial u}\delta u + \left[f(x,u)-\dot{x}\right]^T\delta\lambda dt+o(\|\delta x\|,\|\delta u\|,\|\delta \lambda\|,\|\delta t_f\|)dt\\
   &=  \left[\frac{\partial q_T}{\partial x}+ \lambda_f\frac{\partial \varphi}{\partial x}-\lambda(t_f)\right]\delta x_f+\left[\lambda_f\frac{\partial \varphi}{\partial t} + \frac{\partial q_T}{\partial t}+\mathcal{H}(x,u,\lambda,t_f))\right]\delta t_f\\ &+\int\limits_{0}^{t_f} \left[\dot{\lambda}+\frac{\partial \mathcal{H}}{\partial x}\right]\delta x + \frac{\partial \mathcal{H}}{\partial u}\delta u + \left[f(x,u)-\dot{x}\right]^T\delta\lambda dt+o(\|\delta x\|,\|\delta u\|,\|\delta \lambda\|,\|\delta t_f\|)dt\\
\end{align*}
The fundamental theorem of calculus of variation states that if $(x^\star,u^\star)$ is extrema, then the variations $\delta J$ (linear terms of $\delta x,\delta u,\delta x_f,\delta t_f$) must be zero. Since $\lambda$ can be chosen arbitrarily, we choose $\lambda^\star$ such that the linear terms of $\delta x$ is $0$, i.e.
\begin{equation}
\label{eq:appendix:pmp_costate}
    \dot{\lambda}^\star+\frac{\partial \mathcal{H}}{\partial x}\Big|_\star = 0
\end{equation}
Since the $(x^\star,u^\star)$ must satisfy the constraint in \eqref{eq:appendix:dynamics_constraint},
\begin{equation}
\label{eq:pmp_xu}
    f(x^\star,u^\star) - \dot{x}^\star = 0 
\end{equation}
 If $u$ is unbounded, we consider all perturbations $\delta u$ such that $\delta x_f$ and $\delta t_f$ is $0$. By the fundamental lemma of calculus of variation, its coefficient function must be
zero; thus,
\begin{equation}
\label{eq:appendix:pmp_u}
    \frac{\partial \mathcal{H}}{\partial u}\Big|_\star = 0
\end{equation}
This equation is also called Pontryagin's maximum principle.
The rest of variations are therefore $0$, i.e., 
\begin{equation*}
    \left[\frac{\partial q_T}{\partial x}+ \lambda_f\frac{\partial \varphi}{\partial x}-\lambda(t_f)\right]^T\delta x_f+\left[\lambda_f\frac{\partial \varphi}{\partial t} + \frac{\partial q_T}{\partial t}+\mathcal{H}(x,u,\lambda,t_f))\right]\delta t_f
\end{equation*}
Note that since $\varphi(x_f+\delta x_f,t_f+\delta t_f) = \varphi(x_f,t_f)=0$, we have 
$$\frac{\partial \varphi}{\partial x_f}\delta \gamma_f + \frac{\partial \varphi}{\partial t}\delta t_f = 0 $$
i.e.
$$\begin{bmatrix}
    \frac{\partial \varphi}{\partial x_f} \\ \frac{\partial \varphi}{\partial t}
\end{bmatrix}^T\begin{bmatrix}
    \delta x_f\\ \delta t_f
\end{bmatrix}=0$$
The admissible $\begin{bmatrix}
    \delta x_f\\ \delta t_f
\end{bmatrix}$ are on a hyperplane normal to the vector
$\begin{bmatrix}
    \frac{\partial \varphi}{\partial x_f} \\ \frac{\partial \varphi}{\partial t_f}
\end{bmatrix}$.
Therefore $\begin{bmatrix}
    \frac{\partial q_T}{\partial x}-\lambda(t_f) \\ \frac{\partial q_T}{\partial t} + \mathcal{H}
\end{bmatrix}$ and $\begin{bmatrix}
    \frac{\partial \varphi}{\partial \gamma} \\ \frac{\partial \varphi}{\partial t}
\end{bmatrix}$ are colinear. We can choose $\lambda_f$ such that
\begin{equation}
   \begin{bmatrix}
    \frac{\partial q_T}{\partial x}-\lambda(t_f) \\ \frac{\partial q_T}{\partial t} + \mathcal{H}
\end{bmatrix}=\lambda_f\begin{bmatrix}
    \frac{\partial \varphi}{\partial \gamma} \\ \frac{\partial \varphi}{\partial t}
    \end{bmatrix}
    \label{eq:appendix:pmp_bc}
\end{equation}
We consider two special cases that present in our experiment: 1) the terminal state $x_f$ is fixed, and 2) the terminal time is fixed.

1) First, if the terminal state is fixed and terminal time is free, i.e., $\delta x_f=0$. Then $\delta t_f$ can be arbitrary and coefficients of $\delta t_f$ must be $0$, i.e.,
\begin{equation}
\label{eq:appendix:pmp_bc1}
\begin{split}
    \left[ \frac{\partial q_T}{\partial t}\Big|_{\star,t^\star_f}+\mathcal{H}(x^\star,u^\star,\lambda^\star,t^\star_f)\right]&= 0\\
    x^\star(t^\star_f) = x_f
\end{split}
\end{equation}

2) Now we consider the case where the terminal time is fixed and the terminal state is free, i.e., $\delta t_f=0$. Then $\delta x_f$ can be arbitrary and coefficients of $\delta x_f$ must be $0$, i.e.,
\begin{equation}
\label{eq:appendix:pmp_bc2}
\begin{split}
    \left[\frac{\partial q_T}{\partial x}\Big|_{\star,t^\star_f}-\lambda(t^\star_f)^T\right]=0\\
    t^\star_f = t_f
\end{split}
\end{equation}


The~\eqref{eq:appendix:pmp_bc1} and~\eqref{eq:appendix:pmp_bc2} allow us to determine the optimal terminal state or optimal terminal time when they are free and to be optimized.

\section{Kalman Filtering Derivation}\label{appendix:kalman}
The performance measure for designing optimal linear filter is 
\begin{align*}
    J(\Sigma,G) &= q_T(\Sigma(T),T)\\
    &= \textrm{tr}(\Sigma(T))
\end{align*}
Since the terminal time $t_f=T$ is specified and terminal state is free, ~\eqref{eq:appendix:pmp_bc2} applies. Pontryagin's maximum principle yields
\begin{subequations}
    \label{eq:PMP_kalman_appendix}
    \begin{equation}\label{eq:appendix:pmp_kalman_state}
        \dot{\Sigma}^\star=\Bigl[A-G^\star C\Bigr]\Sigma\! +\Sigma\Big[A\!-G^\star C\Big]^T\!+BQB^T\!+G^\star R{G^\star}^T
    \end{equation}
    \begin{equation}\label{eq:appendix:pmp_kalman_costate}
\displaystyle \frac{\partial \textrm{tr}(\lambda^\star\dot{\Sigma}^\star)}{\partial \Sigma}\Big|_{\star}+\dot{\lambda^{\star}}^T= 0
    \end{equation}
    \begin{equation}
        \label{eq:appendix:pmp_kalman_u}
        \displaystyle \frac{\partial \textrm{tr}(\lambda^\star\dot{\Sigma}^\star)}{\partial G}\Big|_{\star} = 0 
    \end{equation}    
    \begin{equation}\label{eq:appendix:pmp_kalman_bc}
        \lambda^\star(T)^T = \bm I_n
    \end{equation}
    \begin{equation}\label{eq:appendix:pmp_kalman_init}
        \Sigma^\star(0) = \Sigma_0
    \end{equation}
\end{subequations}
Simplifying~\eqref{eq:appendix:pmp_kalman_costate} yields,
\begin{equation}
    \label{eq:appendix:reduce_pmp_kalman_costate}
    \dot{\lambda}^\star=-\lambda^\star\Bigl[A-G^\star C\Bigr]-\Bigl[A-G^\star C\Bigr]^T\lambda^\star
\end{equation}
From~\eqref{eq:appendix:pmp_kalman_bc} and~\eqref{eq:appendix:reduce_pmp_kalman_costate}, we can conclude that $\lambda^\star$ is symmetric positive definite.
Substitute $\dot{\Sigma}^\star$ in~\eqref{eq:appendix:pmp_kalman_u} by R.H.S expression in~\eqref{eq:appendix:pmp_kalman_state} yields,
\begin{equation}
    \label{eq:appendix:solution_kalman}
    2\lambda^\star\Bigl[2G^\star R-2{\Sigma^\star}C^T\Bigr] = 0
\end{equation}
Since $\lambda^\star$ is invertible,
\begin{equation}
    \label{eq:appendix:kalman_solution_u}
    G^\star = {\Sigma^\star}C^TR^{-1}
\end{equation}
Plugging this solution $G^\star$ in~\eqref{eq:appendix:pmp_kalman_state} yields
\begin{equation}
    \label{eq:appendix:riccati_diff}
    \dot{\Sigma}^\star = A\Sigma^\star + \Sigma^\star A^T+BQB-\Sigma^\star C^T R^{-1}C\Sigma^\star
\end{equation}
which is the matrix differential equation of the Riccati type. The solution $\Sigma^\star$ can be derived from the initial condition $\Sigma^\star(0)=\Sigma_0$ and the differential equation~\ref{eq:appendix:riccati_diff}.
\section{Bang-bang control derivation}\label{appendix:bangbang}
Since the terminal state $x_f$ is specified and terminal time is free, ~\eqref{eq:appendix:pmp_bc1} applies. Pontryagin's maximum principle yields

\begin{subequations}\label{eq:bangbang_appendix}
    \begin{equation}\label{eq:bangbang_x}
        \dot{x}^\star = \begin{bmatrix}
            x^\star_2\\0
\end{bmatrix}+\begin{bmatrix}
            0\\u^\star
\end{bmatrix}
    \end{equation}
    \begin{equation}\label{eq:bangbang_p}
\dot{\lambda}^\star= \begin{bmatrix}
        0\\
        -\lambda_1^\star
    \end{bmatrix}\\
    \end{equation}
    \begin{equation}\label{eq:bangbang_u}
        u^\star = \argmin_u  1+\lambda^\star_1x^\star_2 + \lambda^\star_2 u
    \end{equation}
    \begin{equation}\label{eq:bangbang_bc}
          1+\lambda_1^\star(t^\star_f)x^\star_2(t^\star_f) + \lambda^\star_2(t^\star_f) u^\star(t^\star_f) = 0
    \end{equation}
    \begin{equation}\label{eq:bangbang_init}
          \begin{bmatrix}
              x^\star_1(0)\\x^\star_2(0)
          \end{bmatrix} = \begin{bmatrix}
              x_0\\v_0
          \end{bmatrix}
    \end{equation}
    \begin{equation}\label{eq:bangbang_end}
          \begin{bmatrix}
              x^\star_1(t^\star_f)\\x^\star_2(t^\star_f)
          \end{bmatrix} = \begin{bmatrix}
              0\\0
          \end{bmatrix}
    \end{equation}
\end{subequations}
The~\eqref{eq:bangbang_u} yield
\begin{align*}
    \forall u, 1+\lambda_1^\star x_2 + \lambda_2^\star u^\star &\leq 1+\lambda_1^\star x_2 + \lambda_2^\star u\\
    u^\star = \begin{cases}
        -\textrm{sign}(\lambda_2) & \textrm{if $\lambda_2^\star \neq 0$} \\
        \quad\textrm{indeterminate} & \textrm{if $\lambda_2^\star = 0$}
    \end{cases}
\end{align*}
Assuming that $\lambda^\star_2$ is \textbf{not} a zero function, the~\eqref{eq:bangbang_p} yields
\begin{equation}
    \label{eq:general_form_p}
    \begin{split}
    \lambda^\star_1(t) &= c_1\\
    \lambda^\star_2(t) &= -c_1t+c_2\\
    \end{split}
\end{equation}
where $c_1,c_2$ are constants to be determined. We see from~\eqref{eq:general_form_p} that $\lambda_2$ changes sign at most once. There are two possible cases:
\begin{enumerate}
    \item $\lambda^\star_2$ sign remains constant in $[0,t^\star_f]$
    \item $\lambda^\star_2$ changes sign in $[0,t^\star_f]$
\end{enumerate}
 For case 1, we have the general form of 
\begin{equation}
\label{eq:case12}
    \begin{split}
    x_2(t) &=v_0 + at  \quad\qquad\textrm{for $t\in [0,t^\star_f]$}\\
    x_1(t) &= p_0 + v_0t + \displaystyle \frac{1}{2}at^2 \;\ \quad\textrm{for $t\in [0,t^\star_f]$}
    \end{split}
\end{equation}
For case 2, we have the general form of $x$    
\begin{equation}
\label{eq:case34}
    \begin{split}
    x_2(t) &= \begin{cases}
                v_0 + at  & \quad\textrm{if }{t\leq t_{m}}\\
                v_0 + at_{m}-a(t-t_m) & \quad\textrm{if }{ t^\star_f \geq t\geq t_{m}}
            \end{cases}\\
    x_1(t) &= \begin{cases}
                p_0 + v_0t + \displaystyle \frac{1}{2}at^2 & \quad\textrm{if }{t\leq t_{m}}\\
                x_0+v_0t+3att_{m}-2at_m^2-\displaystyle \frac{1}{2}at^2 & \quad\textrm{if }{t\geq t_{m}}\\
            \end{cases}
    \end{split}
\end{equation}
where $a=\pm 1$ and $t_m$ is the time where $\lambda^\star_2$ switches sign.
To determine which case corresponds to the system, we validate with the boundary condition.
Suppose we try with the general expression in~\eqref{eq:case34} and substitute in boundary conditions in~\eqref{eq:bangbang_appendix}:
\begin{equation}
    \label{eq:bangbang_constraint}
    \begin{split}
        -c_1t_m + c_2 &= 0 \quad\quad \textrm{( From the condition $\lambda^\star_2(t_m)=0$)} \\
        a &= \pm 1\\
        v_0+at_m-a(t_f-t_m) &= 0 \quad\quad \textrm{( From \eqref{eq:bangbang_end}})\\
        p_0+v_0t+3at_ft_{m}-2at_m^2-\displaystyle \frac{1}{2}at_f^2 &= 0 \quad\quad \textrm{( From \eqref{eq:bangbang_end}})\\
        1-a(c_1t_f+c_2) &= 0  \quad\quad \textrm{( From \eqref{eq:bangbang_bc}})
    \end{split}
\end{equation}

Specific example: $x_0 = 1, v_0 = 0$.

Solving \eqref{eq:bangbang_constraint} yields
\begin{equation}
\label{eq:bangbang_solution_solution_explicit}
\begin{split}
    t_f &= 2t_m\\
    1 &= -at_m^2\\
    a &= -1\\
    t_m &= 1\\
    c_1 &= 1\\
    c_2 &= 1
\end{split}
\end{equation}
which means the system with initial state condition $(1,0)$ falls the second case. If we substitute the general expression~\eqref{eq:case12} instead, there would be no solutions satisfying~\eqref{eq:bangbang_constraint}.

\textbf{Remarks}: This derivation of bang-bang solution is based on assumption that $\lambda$ is not a zero function.
\section{Geodesics derivation}\label{appendix:geodesics}
Suppose we want to find the curve on a manifold $\mathcal{M}\subset \mathbb{R}^3$ with minimal length starting from point $p_0\in \mathcal M$  to a submanifold (or a boundary) $\mathcal{N}$ in $\mathcal{M}$. The optimization problem can be written as 

\begin{equation}
\begin{aligned}
\min_{\gamma} \quad &L(\gamma) \\
\textrm{s.t.} \quad &\gamma(t) \in \mathcal{M} \\
&\gamma(0)=p_0\\
&\gamma(1)\in\mathcal{N}
\end{aligned}
\label{eq:appendix:length_min}
\end{equation}

First we see that reparameterizing a curve $\gamma$ will not change its arc length. In fact, for any homomorphism $\phi$ from $[0,1]\rightarrow[0,1]$, $\gamma\circ\phi$ has the same trace as $\gamma$ and therefore same arc length. Therefore the minimizer of $L$ is not unique. However, the minimizer of $E(\gamma)$ 
\begin{equation*}
    E(\gamma) = \int\limits_{0}^{1} \|\dot{\gamma}(t)\|^2 dt
\end{equation*}
is unique. The minimizer of $E(\gamma)$ is also the minimizer of $L(\gamma)$. In fact, let $\gamma^\star$ be the minimizer of $E(\gamma)$. By Cauchy-Schwarz inequality
$$L(\gamma)^2\leq E(\gamma)$$ 
with equality when $\|\gamma\|$ is constant. We can choose a reparameterization of $\gamma^\star$ such that the new curve $\gamma_c$ has a constant speed. We see that $E(\gamma_c)=L(\gamma)^2\leq E(\gamma)$, therefore $\gamma=\gamma_c$. Now suppose by contradiction that there is another curve $\gamma_1$ such that $L(\gamma_1)< L(\gamma^\star)$. We reparameterize $\gamma_1$ such that the new $\gamma_2$ has the same length as $\gamma_1$ and we have
 $$E(\gamma_2)=L(\gamma_1)^2 < L(\gamma^\star)^2=E(\gamma^\star)$$, contradiction.

\begin{equation}
\begin{aligned}
\min_{\gamma} \quad &\int\limits_{0}^{1}\|\dot{\gamma}\|^2 \\
\textrm{s.t.} \quad &f(\gamma(t))=0 \\
&\gamma(0)=p_0\\
&\varphi(\gamma(1))=0
\end{aligned}
\label{eq:appendix:energy_min}
\end{equation}
Suppose the stopping set is the intersection of two surfaces by the equation $f(\gamma(1))=0$ (i.e.,$\gamma(1)$ is on a manifold) and another surface $h(\gamma)=0$. Then we can define $\varphi = f^2 + h$. In this case, $\varphi$ vanishes if and only if $f(\gamma)=h(\gamma)=0$.
We form the new functional with lagrangian multipliers
$$\mathcal{J} (x,\lambda,\lambda_f) = \lambda_f\varphi(x_f)+ \int\limits_{0}^{1}\|\dot{\gamma}\|^2 +\lambda(t)^Tf(\gamma(t))dt$$
\begin{equation}
\begin{aligned}
    \delta\mathcal{J} &= \lambda_f\frac{\partial \varphi}{\partial \gamma}\delta\gamma_f+\delta\lambda_f\varphi + \int\limits_{0}^{1} \dot{\gamma}\delta \dot{\gamma} + \delta\lambda^Tf(\gamma)+\lambda^T\frac{\partial f}{\partial \gamma}\delta\gamma dt\\
    &=\lambda_f\frac{\partial \varphi}{\partial \gamma}\delta\gamma_f+\delta\lambda_f\varphi +\dot{\gamma(1)}\delta \gamma_f+\int\limits_{0}^{1} \left[\lambda^T\frac{\partial f}{\partial \gamma}-\ddot{\gamma}\right]\delta\gamma + \delta\lambda^Tf(\gamma) dt\\
    &=\left[\lambda_f\frac{\partial \varphi}{\partial \gamma}+\dot{\gamma}(1)\right]\delta\gamma_f+\delta\lambda_f\varphi +\int\limits_{0}^{1} \left[\lambda^T\frac{\partial f}{\partial \gamma}-\ddot{\gamma}\right]\delta\gamma + \delta\lambda^Tf(\gamma) dt\\
\end{aligned}   
\end{equation}
Since $f(\gamma+\delta\gamma)=f(\gamma)=0$, the admissible variation $\delta\gamma$ is a hyperplane with a normal vector $\frac{\partial f}{\partial \gamma}$, i.e. a tangent plane. Therefore $\langle\ddot{\gamma},\delta\gamma\rangle$ for all tangent vector $\delta\gamma$, hence $\ddot{\gamma}$ is colinear with $\frac{\partial f}{\partial \gamma}$. We can choose $\lambda(t)$ such that $$\lambda(t)\frac{\partial f}{\partial \gamma} =\ddot{\gamma}(t)$$

Next $\delta\gamma_f$ must be perpendicular to each vector $\begin{bmatrix}
    \frac{\partial f}{\partial \gamma}
\end{bmatrix}$ and $\begin{bmatrix}
    \frac{\partial h}{\partial \gamma}
\end{bmatrix}$. Since $\dot{\gamma}(t)$ is perpendicular to $\frac{\partial f}{\partial \gamma}$, we deduce that $\dot{\gamma}$ must be parallel to $\frac{\partial \varphi}{\partial \gamma}$ and we can choose $\lambda_f$ such that
$$\lambda_f\frac{\partial \varphi}{\partial \gamma}=\dot{\gamma}$$

\end{document}